\documentclass[lettersize,journal]{IEEEtran}
\usepackage{amsmath,amsfonts}
\usepackage{algorithmic}
\usepackage{algorithm}
\usepackage{array}
\usepackage[caption=false,font=normalsize,labelfont=sf,textfont=sf]{subfig}
\usepackage{textcomp}
\usepackage{stfloats}
\usepackage{url}
\usepackage{verbatim}
\usepackage{graphicx}
\usepackage{cite}
\usepackage{multirow}
\usepackage{booktabs}
\usepackage{amssymb}
\usepackage[table]{xcolor}   
\usepackage{bm}
\usepackage{float}
\usepackage{pifont}
\usepackage{xcolor}
\usepackage{hyperref}
\providecommand{\cmark}{\ding{51}}
\providecommand{\xmark}{\ding{55}}
\begin{document}

\title{SR-Nav: Spatial Relationships Matter for \\ Zero-shot Object Goal Navigation}

\author{Leyuan Fang,~\IEEEmembership{Senior Member,~IEEE}, Zan Mao, Zijing Wang, and Yinlong Yan
        % <-this % stops a space
\thanks{This work was supported in part by the National Natural Science Foundation of China under the National Science Fund for Distinguished Young Scholars 62425109 and Grant U22B2014, in part by the Fundamental and Interdisciplinary Disciplines Breakthrough Plan of the Ministry of Education of China under Grant JYB2025XDXM122. (Corresponding author: Yinlong Yan.)

Leyuan Fang, Zan Mao, Zijing Wang, and Yinlong Yan are with the School of Artificial Intelligence and Robotics, Hunan University, Changsha, China, 410114, China (e-mail:
fangleyuan@gmail.com; mzzyw0421@163.com; wzj\_183@hnu.edu.cn; yanyl@hnu.edu.cn)
}% <-this % stops a space
\thanks{}}

\markboth{Journal of \LaTeX\ Class Files,~Vol.~14, No.~8, August~2021}%
{Shell \MakeLowercase{\textit{et al.}}: A Sample Article Using IEEEtran.cls for IEEE Journals}


\maketitle

\begin{abstract}
Zero-shot object-goal navigation aims to find target objects in unseen environments using only egocentric observation. Recent methods leverage foundation models' comprehension and reasoning capabilities to enhance navigation performance. However, when faced with poor viewpoints or weak semantic cues, foundation models often fail to support reliable reasoning in both perception and planning, resulting in inefficient or failed navigation. We observe that inherent relationships among objects and regions encode structured scene priors, which help agents infer plausible target locations even under partial observations. Motivated by this insight, we propose Spatial Relation-aware Navigation (SR-Nav), a framework that models both observed and experience-based spatial relationships to enhance both perception and planning. Specifically, SR-Nav first constructs a Dynamic Spatial Relationship Graph (DSRG) that encodes the target-centered spatial relationships through the foundation models and updates dynamically with real-time observations. We then introduce a Relation-aware Matching Module. It utilizes relationship matching instead of naive detection, leveraging diverse relationships in the DSRG to verify and correct errors, enhancing visual perception robustness. Finally, we design a Dynamic Relationship Planning Module to reduce the planning search space by dynamically computing the optimal paths based on the DSRG from the current position, thereby guiding planning and reducing exploration redundancy. Experiments on HM3D show that our method achieves state-of-the-art performance in both success rate and navigation efficiency. The code will be publicly available at \href{https://github.com/Mzyw-1314/SR-Nav}{SR-Nav}.
\end{abstract}


\section{Introduction}
\IEEEPARstart{Z}{ero-shot} object-goal navigation (ZSON) aims to enable an agent to accurately locate and approach a user-specified object (e.g., \textit{“find a toilet”}) in unseen environments without any target-specific training, relying solely on egocentric observations \cite{DBLP:journals/corr/abs-2006-13171}. This capability is essential for real-world robotics, including household service, healthcare assistance, and search-and-rescue or warehouse automation, where agents must generalize to novel environments and execute open-ended instructions. Due to the absence of global map information and the invisibility of the target, the core challenge of ZSON is to leverage limited visual input to infer the target’s location and navigate efficiently in unseen environments.

\begin{figure}[!t]
\centering
\includegraphics[width=1.0\linewidth]{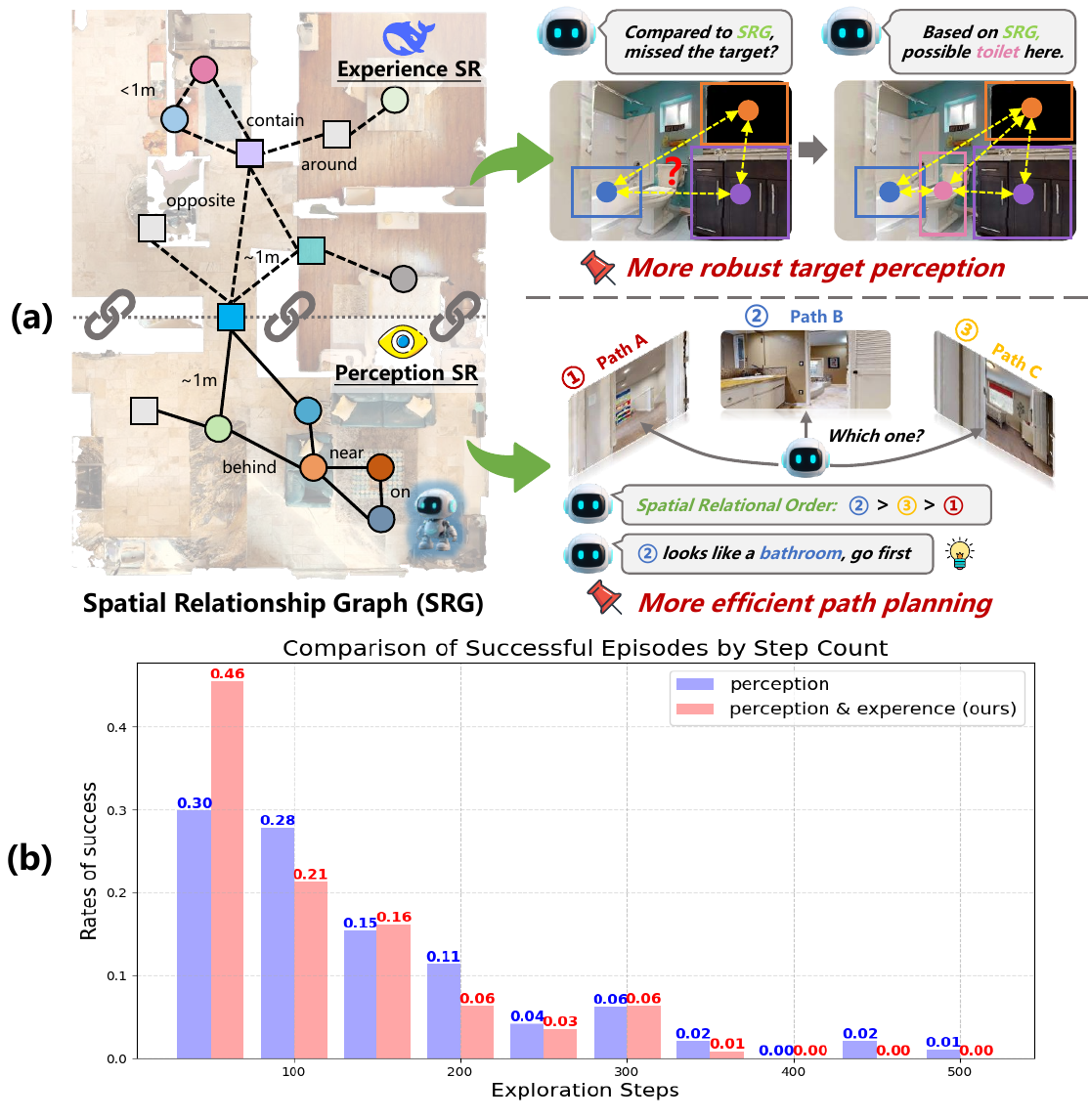}
\caption{(a) Our Spatial Relationship Graph (SRG) integrates experiential spatial priors and perceptual observations to enhance target perception via relational matching and guide efficient path planning through spatial reasoning. (b) Comparison of successful episodes by step count: our perception+experience approach (red) achieves higher success rates at fewer exploration steps than perception-only methods (blue), demonstrating that experiential knowledge provides effective guidance in early, observation-limited navigation.}
\label{fig:motivation}
\end{figure}

\begin{figure*}[t]
\centering
\includegraphics[width=1.0\textwidth]{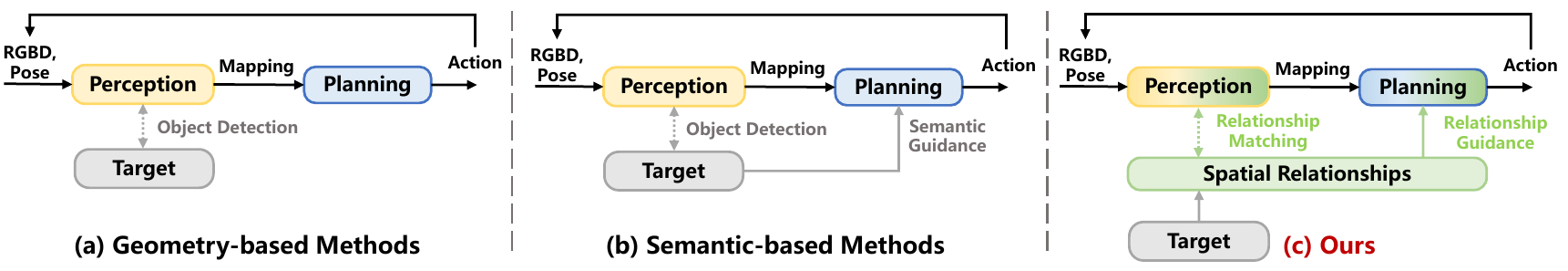} % Reduce the figure size so that it is slightly narrower than the column.
\caption{Comparison with existing methods. (a) Geometry-based methods perceive targets via object detection and solely rely on geometric cues for planning. (b) Semantic-based methods incorporate target semantics to guide planning. (c) Our method employs spatial relationship matching for target perception and utilizes target-associated spatial relationships for planning guidance.}
\label{fig:Comparison}
\end{figure*}

\begin{figure}[t]
  \centering
  % \fbox{\rule{0pt}{2in} \rule{1.0\linewidth}{0pt}}
   \includegraphics[width=1.0\linewidth]{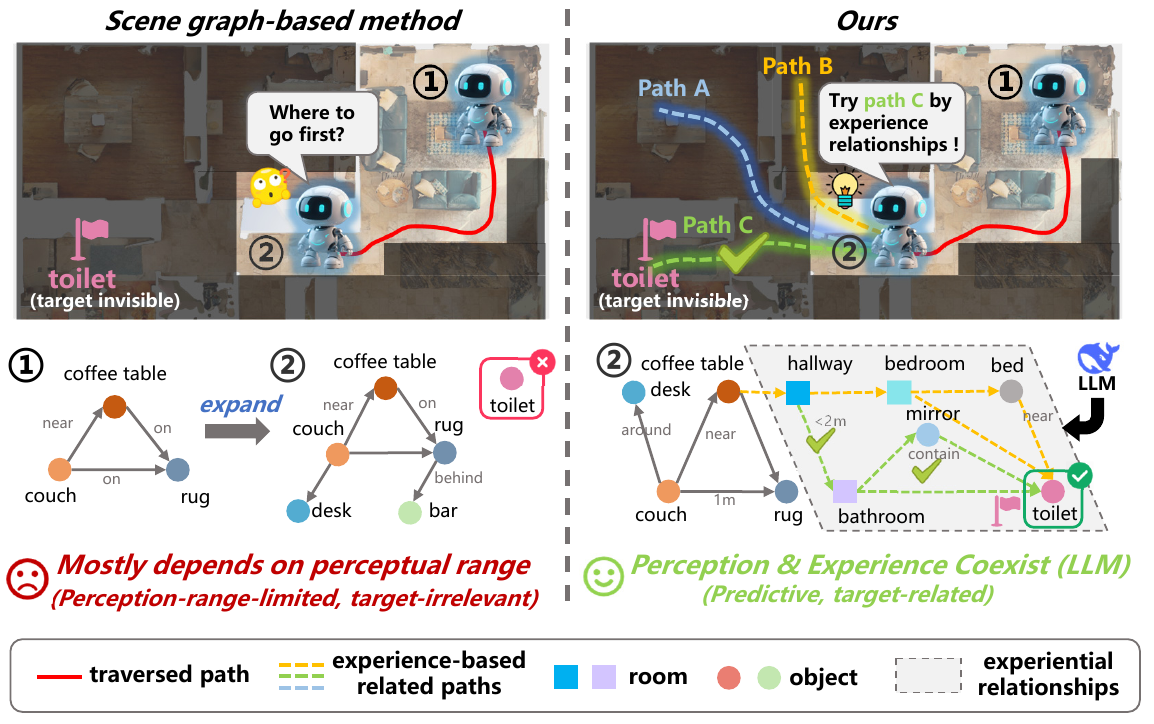}
   \caption{Compared to scene graphs, which are limited by perception and mostly target-irrelevant, our method leverages both observed and experience-based relationships, which are predictive and target-relevant, enabling more effective navigation guidance.}
   \label{fig:Comparison_graph}
\end{figure}

Current ZSON methods can be categorized into two types based on the paradigm used to formulate navigation policies: end-to-end frameworks \cite{cai2024bridging, karnan2022voila, majumdar2022zson, bd7286b950c1479e9a4f8a330722f897, 10161289, wohlke2021hierarchies} and modular frameworks \cite{zhou2023esc, yin2024sg, yokoyama2024vlfm, 10.5555/3692070.3694273, zhong2025_2411.16425}. End-to-end approaches map egocentric observations directly to action sequences through reinforcement or supervised learning. Although certain progress has been achieved, they heavily rely on large-scale annotated data and suffer from limited generalization in complex environments. In contrast, modular approaches decompose navigation into specialized subtasks such as perception and planning. This design effectively reduces overfitting and enhances generalization in unseen environments. Modular approaches can be further categorized into geometry-based and semantics-based methods, as illustrated in Fig. \ref{fig:Comparison}. Geometry-based methods \cite{gadre2023cows, Luo_2024} rely predominantly on metric and spatial representations such as occupancy maps \cite{Elfes2013OccupancyGA}, frontier distributions \cite{yamauchi1997frontier}, or spatial uncertainty \cite{inproceedings} to drive exploration. Their decisions are governed by geometric criteria such as coverage gain, reachability, or map completeness, thereby encouraging rapid coverage of unexplored space, while semantics-based methods \cite{huang2024gamap, yu2023l3mvn, zhong2025_2411.16425} aim to exploit target semantics for efficient exploration. Benefiting from the substantially enhanced reasoning capabilities of recent foundation models \cite{li2023blip, bai2025qwen2, achiam2023gpt}, semantics-based modular methods \cite{zhong2025_2411.16425, yin2024sg, kuang-etal-2024-openfmnav, yu2023l3mvn} now leverage their rich commonsense knowledge to more accurately infer target locations, ultimately enabling more effective navigation. Although empowered by foundation models, existing modular methods still face dual challenges of \textit{perceptual inaccuracies} and \textit{inefficient planning}. Specifically, on the perception side, when targets are distant, occluded, or out of view, egocentric observations often yield incomplete visual features, leading to target misdetection or detection omission, which will result in navigation failure. On the planning side, during long-range exploration, persistently invisible targets and weak semantic cues render existing locally greedy strategies \cite{yokoyama2024vlfm, yin2024sg, huang2024gamap}  ineffective, often producing small and unstable score margins across candidate directions and resulting in exhaustive search and frequent backtracking from suboptimal paths. In summary, insufficient visual and semantic cues limit the reasoning capabilities of foundation models, degrading both perception robustness and planning efficiency. 

We observe that target objects and their relevant objects or regions universally exhibit inherent spatial associations, including orientation, distance, and topological relations (e.g., \textit{''in bathroom, toilet is usually next to a sink within 2 meters''}, see Fig. \ref{fig:motivation} a.). Such Experience-based spatial regularities form universal and intrinsic scene priors that persist even when the target is unobserved. Unlike conventional scene graphs \cite{Im_2024_CVPR, yin2024sg, Werby-RSS-24, chen2024expanding} that only model currently detected entities, these priors integrate empirical knowledge and visual observations, providing robust guidance for perception and planning under limited visual cues. The core of ZSON navigation is to infer target reachability from partial observations, rather than merely verifying target existence. Recent methods \cite{ cao2025cognav, zhou2025beliefmapnav, chaplot2020object} rely on online semantic maps or scene graphs, but are strictly bounded by observable objects and detection reliability. When the target and its correlated cues are invisible, these approaches fail to offer effective reasoning support. To address this gap, as shown in Fig. \ref{fig:Comparison_graph}, our method leverages a target-centered spatial knowledge graph that goes beyond real-time observations. It delivers two key improvements: robust target verification via relational matching, and efficient exploration through relational path reasoning. This enables the agent to perform informed, generalizable navigation beyond current field-of-view, significantly enhancing robustness in unseen environments.

To this end, we propose a \textit{Spatial Relation-aware Navigation framework (SR-Nav)}. We first construct a \textit{Dynamic Spatial Relationship Graph (DSRG)}, where spatial relationships about orientation, distance, and topological relations are generated for the target using carefully designed LLM-based prompting, and dynamically updated according to environmental observations to adapt to the current scene. The DSRG unifies (i) online perceptual relations grounded in current observations and (ii) experience-conditioned spatial priors derived from indoor layout regularities, enabling predictive reasoning beyond immediate visible evidence. To fully exploit the spatial relationships encoded in the DSRG, we introduce two DSRG-driven modules to jointly enhance perception and planning for navigation. The \textit{Dynamic Relationship Planning Module (DRPM)} generates several potential paths to the target based on the DSRG. By localizing the agent within the DSRG using a vision language model (VLM), DRPM generates stepwise guidance to effectively reduce the planning search space (see Fig. 1). After that, the \textit{Relation-Aware Matching Module (RAMM)} is designed to perform spatial relationship matching to enhance the robustness of object perception under poor viewpoints (see Fig. 1). These two modules work synergistically: DRPM guides exploration through relational guidance, while RAMM refines perception through relationship matching, together enabling robust object navigation in complex and unfamiliar environments.

We evaluate our proposed method on several standard benchmarks, including HM3D \cite{DBLP:journals/corr/abs-2109-08238}, MP3D \cite{Matterport3D}, which cover typical challenging scenarios such as long-range targets and severe occlusions. Experimental results show that our SR-Nav achieves higher success rate (SR) and path efficiency (SPL) than existing methods, with notably better generalization in complex environments. Our contributions are summarized as follows
\begin{itemize}
    \item We introduce a target-oriented \emph{Dynamic Spatial Relationship Graph (DSRG)} that dynamically integrates observed spatial evidence with experience-conditioned priors, enabling predictive reasoning over unexplored regions.

    \item We propose a \emph{Relation-Aware Matching Module (RAMM)} that performs relational consistency verification for robust target perception, effectively correcting false positives and recovering missed detections under challenging viewpoints.

    \item We design a \emph{Dynamic Relationship Planning Module (DRPM)} that transforms relational path hypotheses into executable semantic cues, guiding frontier-based exploration toward plausible yet unseen goal-relevant areas beyond observation-bounded reasoning.
\end{itemize}

\section{Related Work}
\label{sec:Related}

\begin{figure*}[t]
  \centering
   \includegraphics[width=1.0\linewidth]{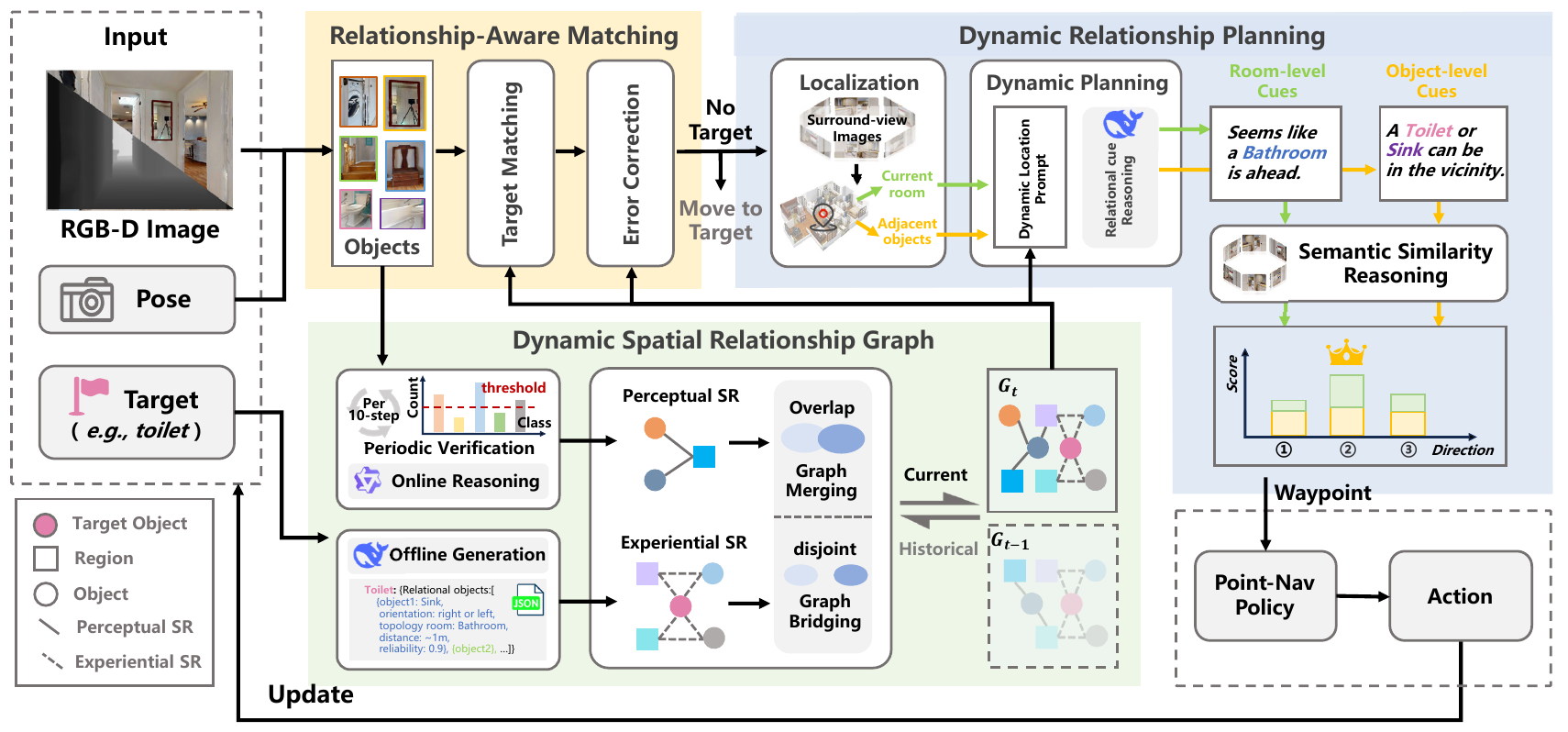}
   \caption{Overview of the SR-Nav. The LLM initializes a Dynamic Spatial Relationships Graph (DSRG) for the target. At each time-step, the agent collects RGB-D observations and updates the DSRG via VLM reasoning. The RAMM module corrects FP/FN errors using spatial priors. When the target is not detected, DRPM integrates DSRG localization and relational cues to generate prompts, ranks scene regions via VLM semantic similarity, and selects the highest-scoring frontier for local navigation.}
   \label{fig:SR-Nav}
\end{figure*}

In this section, we first review the ZSON task and its evolution, followed by a discussion of the application of foundation models to this problem, and finally an overview of how various forms of prior knowledge have been exploited to improve navigation performance.

\textbf{Zero-shot Object Navigation}. ZSON methods are broadly categorized into end-to-end and modular approaches, with our work focusing on the latter for its superior interpretability and flexibility. Recently, frontier-based exploration has emerged as a dominant strategy in ZSON, guiding agents to navigate toward the boundary between known and unknown regions to maximize environmental coverage \cite{yin2024sg, yokoyama2024vlfm, gadre2023cows, huang2024gamap, zhang2024trihelper, inproceedingsfrontier}. As illustrated in Fig. 2, existing frontier-based modular methods can be further divided into geometry-based and semantics-based strategies: geometry-based methods \cite{gadre2023cows, Luo_2024, fuel_zhou} select frontiers purely based on geometric cues (e.g., distance, coverage), aiming to rapidly traverse and cover unknown regions but completely disregard target object semantics, leading to goal-agnostic navigation. in contrast, semantics-based methods leverage target information to prioritize target-relevant frontiers for efficient goal-oriented exploration, which can be subcategorized into structured representation-based and direct embedding-based approaches. 

 Structured representation-based works \cite{yu2023l3mvn, 10285099, zhong2025_2411.16425, kuang-etal-2024-openfmnav, yin2024sg, loo2024open, Hu2021AgentCentricRG} construct structured environmental representations such as scene graphs to encode object categories and spatial relations, leveraging LLMs or VLMs to infer target-relevant frontiers, yet these representations are strictly bounded by real-time observed entities and lack actionable semantic priors when the target is invisible. direct embedding-based methods \cite{huang2024gamap, zhao2025imaginenav, yokoyama2024vlfm, zhou2025beliefmapnav} bypass explicit structured mapping by directly mapping target objects to frontier visual embeddings for greedy decision-making, but they discard global spatial relational logic and fail to leverage prior experiential knowledge about object co-occurrence. Despite these advances, existing semantics-based methods still rely heavily on online perceptual observations without integrating long-term experiential spatial regularities, limiting their performance in early exploration stages where visual information is scarce. A critical gap that our proposed Spatial Relationship Graph (SRG) addresses by fusing perceptual observations with experiential spatial priors to enhance both target perception and path planning efficiency.

% ZSON methods can be broadly classified into end-to-end and modular approaches. In this work, we focus on the modular paradigm. Recently, some ZSON methods \cite{yin2024sg, yokoyama2024vlfm, gadre2023cows, huang2024gamap, zhang2024trihelper, inproceedingsfrontier} employed frontier-based exploration strategy to guide agents to explore unknown regions by following the frontier between known and unknown areas. As illustrated in Fig. 2, existing frontier-based modular methods can be further divided into \textit{geometry-based} and \textit{semantic-based} strategies. Geometry-based methods \cite{gadre2023cows, Luo_2024} select frontiers purely based on geometric criteria, aiming to rapidly traverse and cover unknown regions. However, target semantics are not incorporated into planning. In contrast, semantics-based methods leverage target information to prioritize target-relevant frontiers for efficient goal-oriented exploration. Some works\cite{yu2023l3mvn, zhong2025_2411.16425, kuang-etal-2024-openfmnav, yin2024sg, loo2024open} construct structured environmental representations (e.g., scene graph) for LLM or VLM to infer the frontier or waypoint most relevant to the target semantics. While others \cite{huang2024gamap, zhao2025imaginenav, yokoyama2024vlfm} directly map target objects to frontier visual embeddings, greedly makes decisions based on partial information.

\textbf{Foundation Models Guided Navigation}. The emergence of Large Language Models (LLMs) \cite{devlin-etal-2019-bert, DBLP:journals/corr/abs-2501-12948, achiam2023gpt} and Vision-Language Models (VLMs) \cite{bai2025qwen2, li2023blip} has enabled zero-shot reasoning in object-goal navigation. Recent works leverage LLMs for commonsense reasoning via scene-aware prompts. ESC \cite{zhou2023esc} and SG-Nav \cite{yin2024sg} maintain the information of observed regions and objects, while TopV-Nav, Voronoi, ChatNav and L3MVN \cite{zhong2025_2411.16425, 10.5555/3692070.3694273, yu2023l3mvn, chatnav_2025} construct semantic maps with spatial cues. Pixel-Nav \cite{cai2024bridging} further introduces chain-of-thought prompting for planning. However, LLMs often lack grounding in visual textures. In contrast, VLMs align visual observations with textual prompts via multimodal embeddings, reducing language grounding gaps. Approaches like \cite{huang2024gamap, yokoyama2024vlfm, zhao2025imaginenav} use VLMs to transform long-horizon planning into goal-relevant image selection tasks. To mitigate large models' deficiency in 3D spatial understanding, works such as $E_2$BA and STRIVE\cite{E2BANAV, STRIVE2026} integrate the reasoning capabilities of LLMs with the consideration of spatial geometric costs, bridging the gap between high-level reasoning and low-level spatial constraints and thus improving navigation performance. In conclusion, despite recent progress in applying foundation models to visual navigation, existing methods still struggle with unreliable inference under poor viewpoints and weak semantic cues. To address this limitation, we propose a dynamic spatial relationship graph that encodes spatial relationships as explicit priors to guide foundation models in capturing critical scene information, thereby enabling more informed reasoning chains for reliable target localization.

\textbf{Prior knowledge for Navigation.} Recent studies on object-goal navigation mainly fall into two categories in constructing prior knowledge of target objects to assist agents in inferring target positions: spatial layout knowledge \cite{VSNSP, chaplot2020object, ImagineBeforeGo, Zhang_2021_ICCV} and object correlation knowledge \cite{searchforornavigateto, Layout_based, ramakrishnan2022poni}. Specifically, \cite{VSNSP, chaplot2020object} mine static statistical spatial regularities from external scene datasets, yet such priors are overly generalized and fail to adapt to the characteristics of specific scenes. To address this issue, \cite{ImagineBeforeGo} models the global spatial layout priors of scenes via a self-supervised generative map, while \cite{Zhang_2021_ICCV} enables coarse-to-fine region-level subgoal planning by explicitly modeling a hierarchical graph of scenes and objects. Although the aforementioned methods have progressively refined the modeling granularity of spatial layout priors, they still cannot break free from the dependence on the inherent layout of specific scenes, thus suffering from poor transferability in unseen environments. For object correlation knowledge, \cite{LORGTPVN} learns object relation graphs from training environments, where the modeling of spatial correlations is more consistent with test scenes compared with general static priors. \cite{ramakrishnan2022poni} further infers the potential positions of targets by learning object potential functions; \cite{Layout_based} models the contextual relationships among objects with the Dirichlet-Multinomial distribution; and \cite{searchforornavigateto} proposes an object association-based search thinking mechanism. These methods all provide auxiliary spatial cues for target inference when the target is out of the agent’s field of view. However, previous approaches lack an explicit and unified representation of the global multi-dimensional spatial correlations of targets and only encode simple spatial constraints. As a result, they struggle to provide consistent and effective prior guidance for target verification and exploration planning in the early navigation stage with scarce visual cues. To mitigate these limitations, our work constructs a target-centered Spatial Relationship Graph (SRG). This graph explicitly encodes the inherent multi-dimensional spatial correlations among objects and between objects and regions as generalizable prior knowledge, which furnishes reliable and consistent spatial guidance for foundation model reasoning and frontier-based exploration. In addition, the SRG is continuously updated with observational data collected from unseen environments during navigation to adapt to novel scenes.

\section{Approach}
\label{sec:Approach}

\subsection{Object Goal Navigation}
The ZSON navigation task requires an embodied agent to explore an unknown indoor environment and reach an instance of a target category $T$ (e.g., bed, sofa, toilet) without prior visual examples. Each episode $E_i(P_t, P_i, S_i)$ includes a target position $P_t$, an initial pose $P_i$, and a scene sequence $S_i$. Starting from $P_i$, the agent perceives the environment via RGB-D input and updates its pose at each timestep. It selects actions, such as moving forward (0.25 m), turning left / right (30 °), or stopping, based on observations. Success is defined as issuing STOP within a threshold $d_s$ of the target within 500 steps. 

\subsection{Overview}
As illustrated in Fig. 4, SR-Nav first initializes a target-oriented DSRG using LLM-derived spatial priors, which is continuously refined with observations for environmental adaptability. When neither the target nor its associated objects are detected, The dynamic relationship planning module employs a VLM to integrate the DSRG topology with the agent's current location, generating multiple potential paths to the target, which are then decoded as linguistic guidance for the selection of frontiers. Upon detecting potential targets or associated objects, the relation-aware matching module performs relationship matching with the DSRG to suppress false positives and recover missed detections. The verified target coordinates are subsequently passed to a low-level path planner to guide precise navigation toward the target.

\subsection{Dynamic Spatial Relationship Graph}
The DSRG is a structured spatial knowledge representation, defined as a graph $G(V,E)$, where vertices $V$ represent physical entities such as objects and regions, and edges $E$ encode three types of spatial relations: topological defines spatial connectivity and containment, directional define angular bearings relative to the agent, and distance-based quantify euclidean distances between entities. Existing scene graph methods \cite{yin2024sg, zhong2025_2411.16425, 10.5555/3692070.3694273} are limited by perceptual range and mostly capture target-irrelevant knowledge (Fig. \ref{fig:Comparison_graph}). This motivates us to build a representation that combines perception and experience to encode spatial associations. To address this limitation, we employ the DSRG as a persistent and dynamically updated spatial prior. Semantic relation triplets are initialized from LLM-derived commonsense priors and are incrementally updated using real-time observations, thereby maintaining spatial-semantic consistency and enabling robust long-horizon planning.

% \begin{figure}[t]
% \centering
% \includegraphics[width=1.0\columnwidth]{images/figure5.pdf} % Reduce the figure size so that it is slightly narrower than the column. Don't use precise values for figure width.This setup will avoid overfull boxes.
% \caption{Dynamic Relationship Planning Module. This module integrates the agent's location in the DSRG with target-related relations to generate navigation prompts. When the target is absent, it scores frontiers using VLM-based scene-prompt similarity and selects the most promising waypoint.}
% \label{DRPM}
% \end{figure}

\begin{algorithm}[t]
\caption{Graph Refinement}
\label{alg:core_vlm_fusion_fixed}
\begin{algorithmic}[1]

\REQUIRE Prior graph $G_P$, object $o_i$ belong to detections $\{D_t\}_{t=1}^{10}$, $R$ is room type,
\ENSURE $\widehat{G}_P$
\STATE $\widehat{G}_P \gets G_P,\; G_S \gets \varnothing$
    % --- node confirmation ---
\FOR{$o\in D_t$}
    \STATE $C_t \gets $ calculate temporal consistency score (1-4)
    \STATE \textbf{if {$C_t>\eta_{\text{add}}$} then $G_S \gets G_S \cup \{o\}$ endif}
\ENDFOR
%     % --- periodic VLM trigger ---
% \STATE \textbf{if} $10-t > 0$ \textbf{then continue}
\IF{$10-t > 0$}
    \STATE $t \gets t+1$, continue
\ELSE
    % --- relation inference & fusion ---
    \FOR{$o\in G_S$}
        \STATE $R_o, R^\star \gets$ rooms of $o$ and $o^\star$
        \IF{$R_o = R^\star$}   % same-room: object-object
            \STATE $r_{target} \gets $ calculated using Equation (5) 
            % \STATE $c_o^{new} = \lambda c_o^{prior} + (1-\lambda)\alpha$
            \STATE \textbf{if} $(o,o^\star)\in\mathcal{E}_P$ \textbf{then} update
            \STATE $c_o^{new} \gets \lambda c_o^{prior} + (1-\lambda) c_o^{new}$
            \STATE \textbf{else} insert $o-o^\star \gets (r_{target},c_o^{new})$ \textbf{end if}
        \ELSE                 % different-room: room-room
            \STATE $r_{room} \gets $ calculated using Equation (6)
            % \STATE $c_r^{new} = \lambda c_r^{prior} + (1-\lambda)\alpha$
            \STATE \textbf{if} $(r_o,r^\star)\in\mathcal{E}_P$ \textbf{then} update same as above 
            \STATE \textbf{else} insert $r_o-r^\star\gets (r_{room},c_r^{new})$ \textbf{end if}
        \ENDIF
    \ENDFOR
\ENDIF
\STATE \textbf{return} $\widehat{G}_P$

\end{algorithmic}
\end{algorithm}

\textbf{Graph Initialize.} Prior to task execution, we initialize a target-oriented DSRG $G_t(V_t, E_t)$ using common-sense spatial knowledge derived from Large Language Models (LLMs). This initialization is performed offline and the resulting graph is stored as a JSON file for subsequent querying. Specifically, DeepSeek-R1 is prompted under a predefined system configuration \textit{``You are a Spatial Commonsense Reasoner specializing in indoor environments. Your task is to generate physically plausible prior knowledge about spatial relationships using cognitive principles of spatial memory."} to generate three types of spatial relationships via:\textit{`` for the target target-object, please provide typical topological, directional and distance relationships".} This initialization exhibits three desirable properties: target-centricity, multi-relational completeness, and physical plausibility. By explicitly organizing commonsense priors into topological, directional, and distance relations around the target, the initialized graph provides a compact yet sufficiently expressive hypothesis space for subsequent online reasoning, rather than relying on observation-bounded and target-irrelevant associations. Moreover, performing this process offline is methodologically sound, since it decouples high-cost commonsense elicitation from online decision-making while preserving structurally stable indoor regularities.

\begin{figure}[t]
\centering
\includegraphics[width=1.0\columnwidth]{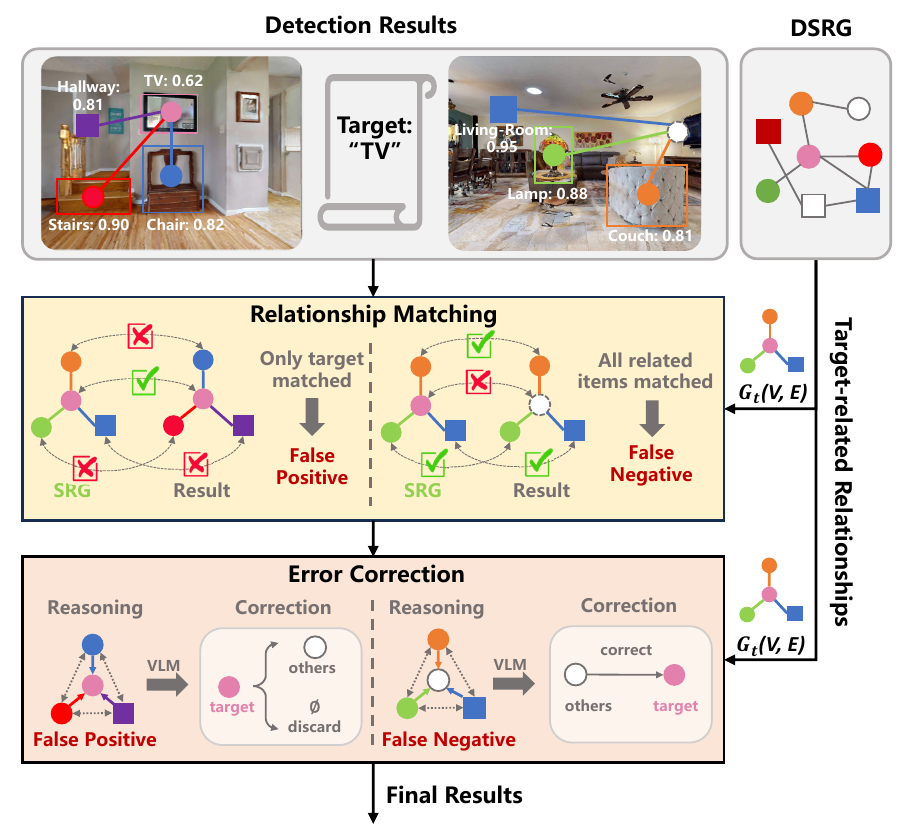} % Reduce the figure size so that it is slightly narrower than the column. Don't use precise values for figure width.This setup will avoid overfull boxes.
\caption{Relationship-aware Matching Module. This module refines raw detections using target-related spatial priors from the DSRG. It uses relationship matching to suppress false positives and identify potential false negatives, ensuring more reliable detection results.}
\label{RAMM}
\end{figure}

\textbf{Graph Refinement.} During exploration, the DSRG is progressively refined by grounding the initialized commonsense prior in temporally accumulated scene evidence. Specifically, the agent first extracts candidate object instances from egocentric observations using open-vocabulary perception, and then performs periodic VLM-based relational reasoning to identify the spatial relations that are both observation-supported and target-relevant. The resulting nodes and informative edges are subsequently fused into the global graph, such that the DSRG evolves from a prior-driven hypothesis space into an observation-aligned spatial representation that increasingly reflects the true object layout of the current environment. For clarity and reproducibility, we summarize the above hierarchical relation inference and confidence-aware graph fusion procedure in Algorithm \ref{alg:core_vlm_fusion_fixed}.

\begin{figure*}[t]
\centering
\includegraphics[width=1.0\textwidth]{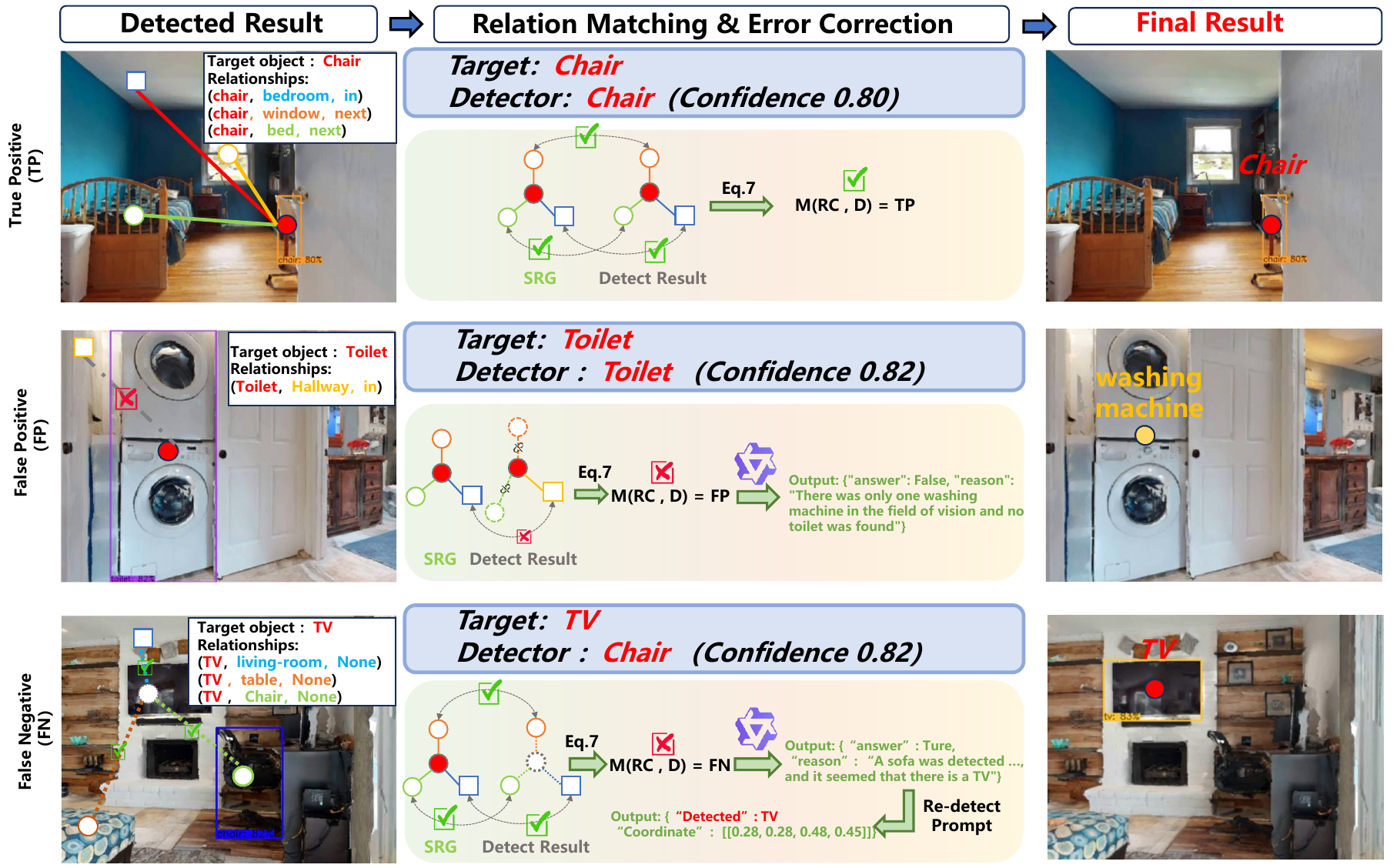} 
\captionsetup{justification=centering}
\caption{Examples of RAMM for perception.}
\label{fig:RAMM}
\end{figure*}

% Node Verification. Single-frame detections are noisy. Therefore, only temporally consistent multi-frame observations should be treated as trustworthy. To this end, we maintain a temporal detection accumulator for each object category. Rather than requiring strict detection in $K$ consecutive frames, we treat a candidate detection as valid if its temporal consistency score within a sliding window of length $K$ exceeds a threshold $\tau$. This formulation is more robust to intermittent occlusion and short-term missed detections while still enforcing temporal stability ($K$ is selected within $[3,8]$ depending on the object type). If the temporal consistency score within the $K$ frames window fails to exceed the threshold, the candidate node is discarded and no instance is added to the dynamic graph.

Node Verification. Single-frame detections are noisy. Therefore, only temporally consistent multi-frame observations should be treated as trustworthy. To this end, we maintain a temporal detection accumulator for each object category. Rather than requiring strict detection in $K$ consecutive frames, we treat a candidate detection as valid if its temporal consistency score within a sliding window of length $K$ exceeds a threshold $\tau$. This formulation is more robust to intermittent occlusion and short-term missed detections while still enforcing temporal stability. Meanwhile, we propose an instance-level verification mechanism that abandons category-wise accumulation. Therefore, we maintain an instance-level existence confidence for each candidate object track \(v_j\). Let \(m_t^{(j)} \in \{0, 1\}\) denote whether a detection in frame \(t\) is associated with \(v_j\), \(c_t^{(j)} \in [0, 1]\) the detector confidence, \(q_t^{(j)} \in [0, 1]\) the observation quality, and \(\nu_t^{(j)} \in [0, 1]\) the expected visibility of \(v_j\) if it exists. We first compute the frame-level existence evidence by integrating positive detection cues and negative non-detection cues:
\begin{equation}
e_t^{(j)} = m_t^{(j)} c_t^{(j)} q_t^{(j)} - \beta \left(1 - m_t^{(j)}\right) \nu_t^{(j)},
\end{equation}
where $\beta$ is a hyperparameter balancing the contribution of negative evidence (non-detection) against positive evidence (detection), set to 0.6 in our experiments to avoid over-penalizing intermittent occlusions. And recursively accumulate it via:
\begin{equation}
s_t^{(j)} = \rho s_{t-1}^{(j)} + e_t^{(j)},
\end{equation}
where $\rho \in (0, 1)$ is the forgetting factor (set to $0.80$) that discounts the weight of past evidence, ensuring the accumulator prioritizes recent observations. To explicitly encourage temporal continuity, we define:
\begin{equation}
r_t^{(j)} = \frac{\sum_{i=1}^{L-1} \lambda^{i-1} m_{t-i+1}^{(j)} m_{t-i}^{(j)}}{\sum_{i=1}^{L-1} \lambda^{i-1} + \epsilon},
\end{equation}
where $L$ is the continuity evaluation window length (set to $5$), $\lambda \in (0, 1)$ is the temporal weight decay factor (set to $0.8$) that emphasizes recent consecutive detections, and $\epsilon = 10^{-6}$ avoids division by zero. This term rewards tracks with sustained detections and penalizes sporadic, discontinuous ones.
The final existence confidence is computed as:
\begin{equation}
C_t^{(j)} = \sigma\left(\alpha s_t^{(j)} + \gamma r_t^{(j)}\right),
\end{equation}
A node is inserted into the DSRG when \(C_t^{(j)} > \eta_{\text{add}}\), and is suppressed, where \(\eta_{\text{add}} = 0.8\) provides hysteresis for stable graph updates.

% Formally, let the most recent $L$ observations of an object be $\{\omega_i\}_{i=0}^{L-1}$, where 
% \begin{equation}
% \omega_i = c_{t-i} \cdot \mathbf{1},
% \end{equation}
% and $c_{t-i}$ denotes the detector confidence in frame $t-i$.  
% The continuous detection strength is accumulated as
% \begin{equation}
% N_{\mathrm{continuous}}^{\omega} = \sum_{i=0}^{L-1} \omega_i ,
% \end{equation}
% which reflects the amount of temporally consistent evidence supporting the existence of the object instance.

% The temporal consistency score is then computed using an exponential smoothing function:
% \begin{equation}
% C_{\mathrm{temporal}} = 1 - \exp\!\left(-\frac{N_{\mathrm{continuous}}^{\omega}}{\tau}\right),
% \qquad \tau = \frac{K}{2},
% \end{equation}
% where $\tau$ controls the rate at which confidence increases. A node is declared as ``existing'' once $C_{\mathrm{temporal}}$ exceeds a predefined threshold. 

VLM performs relational reasoning. Only after a node is confirmed to exist do we infer its semantic relationships: the edges connected to this node are obtained through a Vision-Language Model (Qwen2.5VL-7B), which jointly considers the experience-based spatial priors and the accumulated scene observations. The model predicts two triple relation triples:

(1) Inference of the object-target relationship, if the prior graph contains edges that involve the current room but no edge between the object node and the target node, we directly infer three spatial relations between them:
\begin{equation}
(r_{\mathrm{target}},c_o^{new}) = VLM\left(DSRG, \mathcal{P}_{\mathrm{target}}, \mathcal{P}_{\mathrm{object}}\right),
\end{equation}
where $r_{\mathrm{target}}$ denotes the inferred spatial relation between the target object and another observed object, $c_o^{new}$ denotes the confidence score. $\mathcal{P}_{\mathrm{target}}$ and $\mathcal{P}_{\mathrm{object}}$denote cue words from the target and object.

(2) Room-level indexing and relation inference, if the inferred relations do not connect to prior, we index one level upward, we obtain inter-room relations:
\begin{equation}
(r_{\mathrm{room}},c_r^{new}) = VLM\left(DSRG, \mathcal{P}_{\mathrm{t_{room}}}, \mathcal{P}_{\mathrm{o_{room}}}\right),
\end{equation}
 where $r_{\mathrm{room}}$ denotes the inferred spatial relation between the target room and object room, $c_r^{new}$ denotes the confidence score. $\mathcal{P}_{\mathrm{t_{room}}}$ and $\mathcal{P}_{\mathrm{o_{room}}}$denote cue words from the target room and object room.

After relation inference, we integrate new nodes/edges into the prior graph using a confidence-weighted update rule. 

\begin{figure*}[t]
\centering
\includegraphics[width=1.0\textwidth]{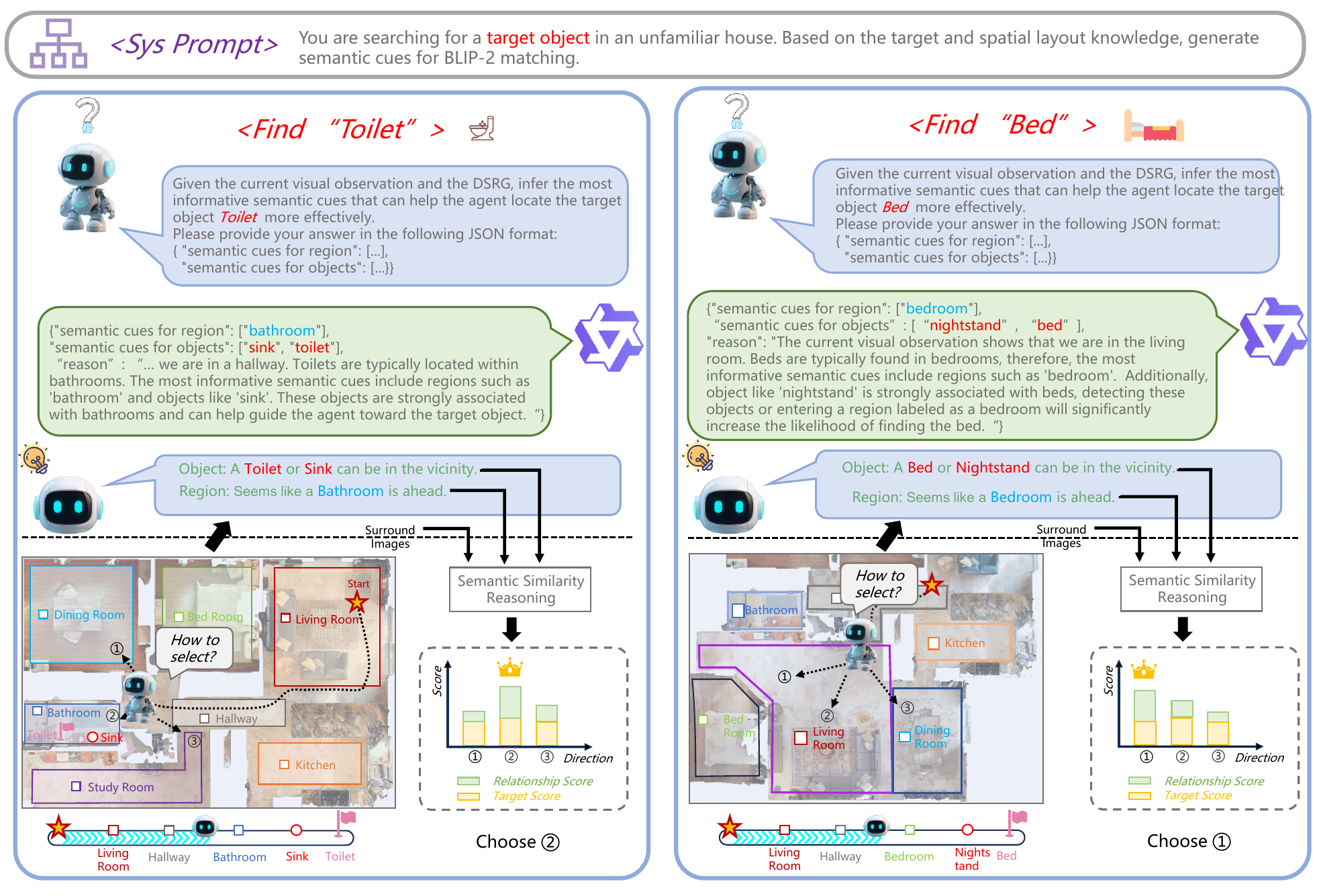} 
\captionsetup{justification=centering}
\caption{Examples of DRPM for planning.}
\label{fig:DRPMa}
\end{figure*}

\subsection{Relation-aware Matching Module (RAMM)}
Successful navigation in ZSON is critically contingent upon the reliability of target detection, yet conventional detectors frequently yield high-frequency false positives or false negatives under occlusion or extreme viewing angles. To address this fundamental limitation, we propose the relation-aware matching module (see Fig. \ref{RAMM}), which formulates target detection as a structured spatial-relation matching process. By jointly optimizing visual-semantic correspondence and spatial relationship constraints, RAMM mitigates perceptual degradation—providing a robust alternative to confidence-based fusion methods while enabling subsequent cross-modal relation refinement and error-corrective graph updating to resolve persistent ambiguities.

\textbf{Relation Matching.} Upon detection of any target instance $d_t$ or other object $d_i$ by the detector, the agent must first verify whether the detection is correct or if any target has been overlooked in the scene. To achieve this, we perform structured spatial matching via the DSRG, formalized as the matching function $ \mathcal{M}(R_C, D) $:
\begin{equation}
\mathcal{M}(R_C, D) = 
\begin{cases} 
\text{TP}, & R_C \in G_t \land \exists\, d_i \in G_t \\
\text{FP}, & R_C \notin G_t \lor \forall\, d_i \notin G_t \\
\text{FN}, & R_C \in G_t \land \exists\, d_i \in G_t \land \nexists\, d_t,
\end{cases}
\end{equation}
where $G_t$ denotes the dynamic spatial relationship graph, and $D$ represents the set of detected object instances (i.e., both $d_i$ and $d_t$ belong to $D$), $R_C$ represents the current region.

The outcome of the detection validation is determined by $ \mathcal{M}(R_C, D) $ the agent assesses the validity of each detection. If $ \mathcal{M}(R_C, D) $ flags an error result, it further classifies the error as a false positive or false negative. This qualitative error diagnosis enables the subsequent error correction stage to take advantage of the VLM for verification and refinement of the detection result.

\textbf{Error Correction.} When potential false positives (FP) are identified, the agent initiates VLM verification using a prompts combining: Detected object visual crop, surrounding context and DSRG relational constraints (e.g., \textit{distinguishing a coffee table from a bed in a living room}). Conversely, when a potential false negative is detected, the agent uses a \textit{``Determine whether the target-object appears in the scene. If found, provide its bounding box coordinates...”} to driven the VLM for descriptive re‑detection of the target $d_t$ (e.g., \textit{locating a chair occluded by a dining table in a restaurant}):
\begin{equation}
d_{\mathrm{t}} = {detectVLM} \bigl( I_{\mathrm{RGB}}, \mathcal{P}_{\mathrm{prompt}}, DSRG \bigr),
\end{equation}
where $I_\mathrm{RGB}$ denotes the current iamge, $P_\mathrm{prompt}$ represents the describe detection prompt.

\subsection{Dynamic Relationship Planning Module (DRPM)}
Efficient agent navigation in unknown environments critically depends on the agent’s ability to plan paths toward unseen targets using scene-level information. While existing methods \cite{yin2024sg, zhong2025_2411.16425, 10.5555/3692070.3694273} enhance foundation model inference by constructing various scene representations to predict potential target locations, they often overlook the rich prior knowledge embedded in the intrinsic spatial relationships of the target. This omission leads to inefficient and unreliable reasoning, especially in complex environments or when semantic cues are weak, making it difficult for the agent to attend to critical information. To address this, we explicitly exploit spatial relationship captured in the DSRG (Fig. \ref{fig:Comparison_graph}). During navigation, the agent is localized within the DSRG in real time. VLMs are then employed to infer potential paths to the target and generate scene-specific guidance cues that highlight key informative regions. Based on this guidance, the agent selects optimal frontiers for exploration. This framework effectively constrains the search space in unknown environments, thereby improving both the efficiency and robustness of navigation.

\definecolor{mygray}{gray}{0.9}
\begin{table*}[t]
\caption{Zero-shot object navigation results on HM3D v0.1 and MP3D benchmarks.}
\label{tab:zson-benchmark}
\centering
\setlength{\tabcolsep}{4mm}
\begin{tabular}{c|c|cc||ccccc}
\toprule
\multirow{2}{*}{\textbf{Method}} & \multirow{2}{*}{\textbf{Venue}} & \multirow{2}{*}{\textbf{Unsupervised}} & \multirow{2}{*}{\textbf{Zero-shot}} & \multicolumn{2}{c}{\textbf{HM3D}} & \multicolumn{2}{c}{\textbf{MP3D}} \\
       &       &              &           & SR $\uparrow$ & SPL $\uparrow$ & SR $\uparrow$ & SPL $\uparrow$ \\
\midrule
SemExp     & NeurIPS'20 & \xmark & \xmark & 37.9 & 18.8 & 36.0 & 14.4 \\
PONI       & CVPR'22    & \xmark & \xmark & -    & -    & 31.8 & 12.1 \\
SGM        & CVPR'24    & \xmark & \xmark & 60.2 & 30.8 & 37.7 & 14.7 \\
\midrule
ZSON       & NeurIPS'22 & \xmark & \cmark & 25.5 & 12.6 & 15.3 & 4.8 \\
PixNav     & ICRA'24    & \xmark & \cmark & 37.9 & 20.5 & -    & - \\
\midrule
ESC        & CVPR'22    & \cmark & \cmark & 39.2 & 22.3 & 28.7 & 14.2 \\
L3MVN      & IROS'23    & \cmark & \cmark & 50.4 & 23.1 & 34.9 & 14.5 \\
SemUtil    & RSS'23     & \cmark & \cmark & 54.0 & 24.9 & -    & - \\
VoroNav    & ICML'24    & \cmark & \cmark & 42.0 & 26.0 & -    & - \\
VLFM       & ICRA'24    & \cmark & \cmark & 52.5 & 30.4 & 34.6 & 16.2 \\
GAMap      & NeurIPS'24 & \cmark & \cmark & 53.1 & 26.0 & -    & - \\
OpenFMnav  & NAACL'24   & \cmark & \cmark & 52.5 & 24.1 & 37.2 & 15.7 \\
SG-Nav     & NeurIPS'24 & \cmark & \cmark & 53.9 & 24.8 & 40.1 & 16.0 \\
Uni-Nav    & CVPR'25    & \cmark & \cmark & 54.5 & 25.1 & \textbf{41.0} & 16.4 \\
ImagineNav & ICLR'25    & \cmark & \cmark & 53.0 & 23.8 & -    & - \\ 
WMNav-Gemini & IROS'25    & \cmark & \cmark & 58.1 & 31.2 & 45.4    & 17.2 \\
WMNav-Qwen2.5vl & IROS'25    & \cmark & \cmark & 46.1 & 20.7 & -    & - \\
\rowcolor{gray!30}
\textbf{SR-Nav}     & \textbf{Ours}       & \cmark & \cmark & \textbf{58.3} & \textbf{33.0} & 37.7 & \textbf{17.1} \\
\bottomrule
\end{tabular}
\end{table*}

\textbf{Dynamic Guidance Generation}. We employ VLM as a dynamic guidance generator, driven by a carefully designed guidance prompt: \textit{``You are a Spatial Path Inference Engine for embodied agents. You are searching for a target object in an unfamiliar house. Based on the target and spatial layout knowledge, generate semantic cues for BLIP-2 matching."}. The framework first localizes the agent within the DSRG using visual context (\textit{Prompt: ``Correlate visual features with DSRG nodes to determine agent's precise graph position"}). It then exploits relational priors in the DSRG, together with the agent’s current position, to infer potential paths toward the target and extract critical guidance cues along these paths (\textit{Prompt: ``Given the current visual observation and the DSRG, infer the most informative semantic cues that can help the agent locate the target object more effectively."}). These cues are subsequently encoded into targeted prompts (\textit{``Seems like a Region-cues is ahead"} and \textit{``A object-cues can be in the vicinity"}) that guide the local planning VLM, enabling more efficient scene parsing through prioritized attention to task-relevant features.

\begin{figure*}[t]
\centering
\includegraphics[width=1.0\textwidth]{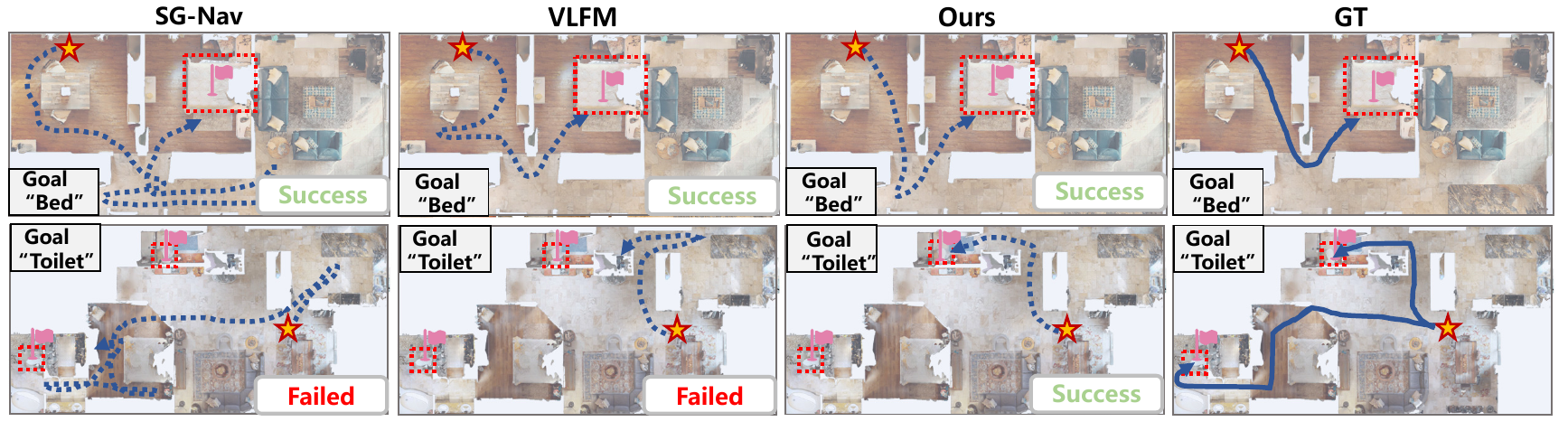} % Reduce the figure size so that it is slightly narrower than the column.
\caption{Qualitative comparison of navigation trajectories on the HM3D v0.1 dataset. Rows 1 and 2 demonstrate higher navigational efficiency and overall success rate of our method, respectively.}
\label{fig:VIS}
\end{figure*}

\textbf{Frontier-Based Exploration with Semantic Guidance}. To generate candidate frontiers $S_t$, we first build a top-down 2D obstacle map using the agent’s current depth and odometry observations. Each pixel is classified into free, occupied, or unknown space, and the map is updated as the agent explores. Following \cite{yamauchi1997frontier}, we identify frontier points as the midpoints between explored and unexplored regions. Given the guidance prompt (see Dynamic Guidance Generation section) and the frontier candidates, we compute the cosine similarity between the prompt and the current visual observation using a VLM. The above procedure can be formalized as:
\begin{equation}
s_t^* = \arg\max_{s_i \in \mathcal{S}_t} \operatorname{Sim}\left(I_\mathrm{RGB-i}, \operatorname{P_{\mathrm{Gen}}}(\operatorname{VLM}(r_i, p_t)) \right),
\end{equation}
where $S_t$ denotes the set of frontier candidates at timestep $t$. The agent's observation of $s_i$ is represented as $I_\mathrm{RGB-i}$ , and its current position related to DSRG is denoted by $P_t$. A relational path $r_i$ is sampled from the DSRG, encoding the spatial connections between the target and related objects. The function \textit{$P_{\mathrm{Gen}}$} transforms the relational path and the agent's current position into a natural language prompt. The function \textit{Sim} computes the semantic similarity between the visual observation and the relational prompt using a VLM, and confidence computation for view overlap follows the approach used in \cite{yokoyama2024vlfm}. The selected frontier $s_t^*$ corresponds to the one with the highest semantic similarity score and is thus chosen as the next navigation waypoint. Subsequently, we adopt the local point navigation planner from VLFM to generate actions toward the given goal.

\section{Experiments}
\label{sec:Experiments}

% \begin{figure*}[t]
% \centering
% \includegraphics[width=1.0\textwidth]{images/fig6.pdf} % Reduce the figure size so that it is slightly narrower than the column.
% \caption{Qualitative comparison of navigation trajectories on the HM3D v0.1 dataset. Rows 1 and 2 demonstrate higher navigational efficiency and overall success rate of our method, respectively.}
% \label{fig:VIS}
% \end{figure*}

\begin{figure*}[t]
\centering
\includegraphics[width=1.0\textwidth]{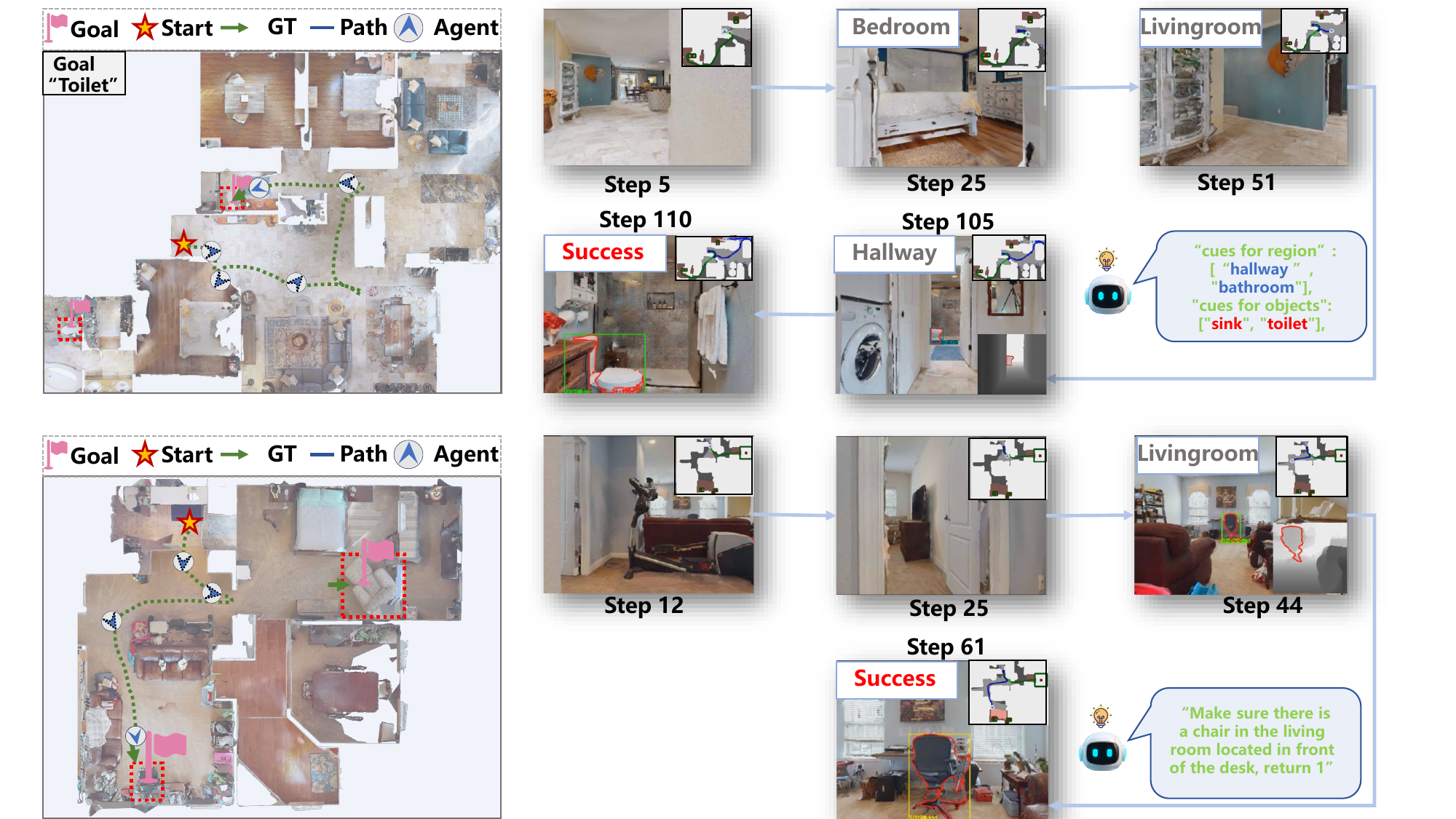} % Reduce the figure size so that it is slightly narrower than the column.
\caption{Visualization of the navigation process of SR-Nav.}
\label{VIS:process}
\end{figure*}

\subsection{Experimental Setup}
\textbf{Datasets and evaluation metric}. We conduct experiments on two widely used benchmarks: Habitat-Matterport3D (HM3D) \cite{DBLP:journals/corr/abs-2109-08238} and Matterport3D (MP3D) \cite{Matterport3D}. HM3D contains 2,000 test episodes sampled from 20 high-fidelity indoor environments spanning 6 object categories, while MP3D includes 2,195 episodes from 11 diverse scenes covering 21 categories. Following standard ZSON evaluation protocols, we adopt Success Rate (SR), which measures the proportion of episodes where the agent successfully reaches the target, and Success weighted by Path Length (SPL), which additionally accounts for the efficiency of the navigation path.

\textbf{Implementation Details}. All experiments are conducted in the Habitat simulator. At each timestep, the agent receives a 480×640 RGB-D image. The valid depth range is [0.5m, 5m], the onboard camera is mounted at a height of 0.88 meters, with a field of view of 79°. The agent can either move forward 0.25 meters or rotate by 30° per step. We use DeepSeek-VL R1 to generate the initial spatial relation graph. For the VLM, we adopt Qwen2.5-VL (7B). For object detection, YOLOv7 \cite{wang2023yolov7} is used for COCO-class targets, and Grounding DINO\cite{liu2024grounding} is used for open-vocabulary detection. SAM\cite{zhang2023faster} is employed to extract object-level 3D information from RGB-D input. All experiments are conducted on two RTX 3090 GPUs with 24GB memory.

% \begin{table*}[t]
% \setlength{\tabcolsep}{4mm}
% \centering
% \caption{Performance comparison on object-goal navigation across six target categories and average. Best results are in \textbf{bold}. We manually verified and annotated the corresponding results. Entries based on manual annotation are marked with an asterisk~($^\ast$).}
% \label{tab:detector}
% \begin{tabular}{lcccccccc}
% \toprule
% \textbf{Method} & \textbf{bed} & \textbf{chair} & \textbf{plant} & \textbf{sofa} & \textbf{toilet} & \textbf{tv\_monitor} & \textbf{Average} \\
% \midrule
% L3MVN      & 52.9 & 51.6 & \textbf{46.4} & 50.1 & 41.5 & \textbf{54.2} & 49.5 \\
% VLFM    & 50.1 & 66.6 & 44.0 & 48.1 & 58.5 & 34.9 & 52.5 \\
% \textbf{SR-Nav (ours)}  & \textbf{55.4} & \textbf{68.9} & 42.8 & \textbf{54.1} & \textbf{65.3} & 44.8 & \textbf{58.3} \\
% \textbf{SR-Nav (ours)$^{*}$}  & - & - & - & - & - & - & \textbf{61.3}$^{*}$ \\
% \bottomrule
% \end{tabular}
% \end{table*}

\begin{figure}[t]
\centering
\includegraphics[width=0.9\columnwidth]{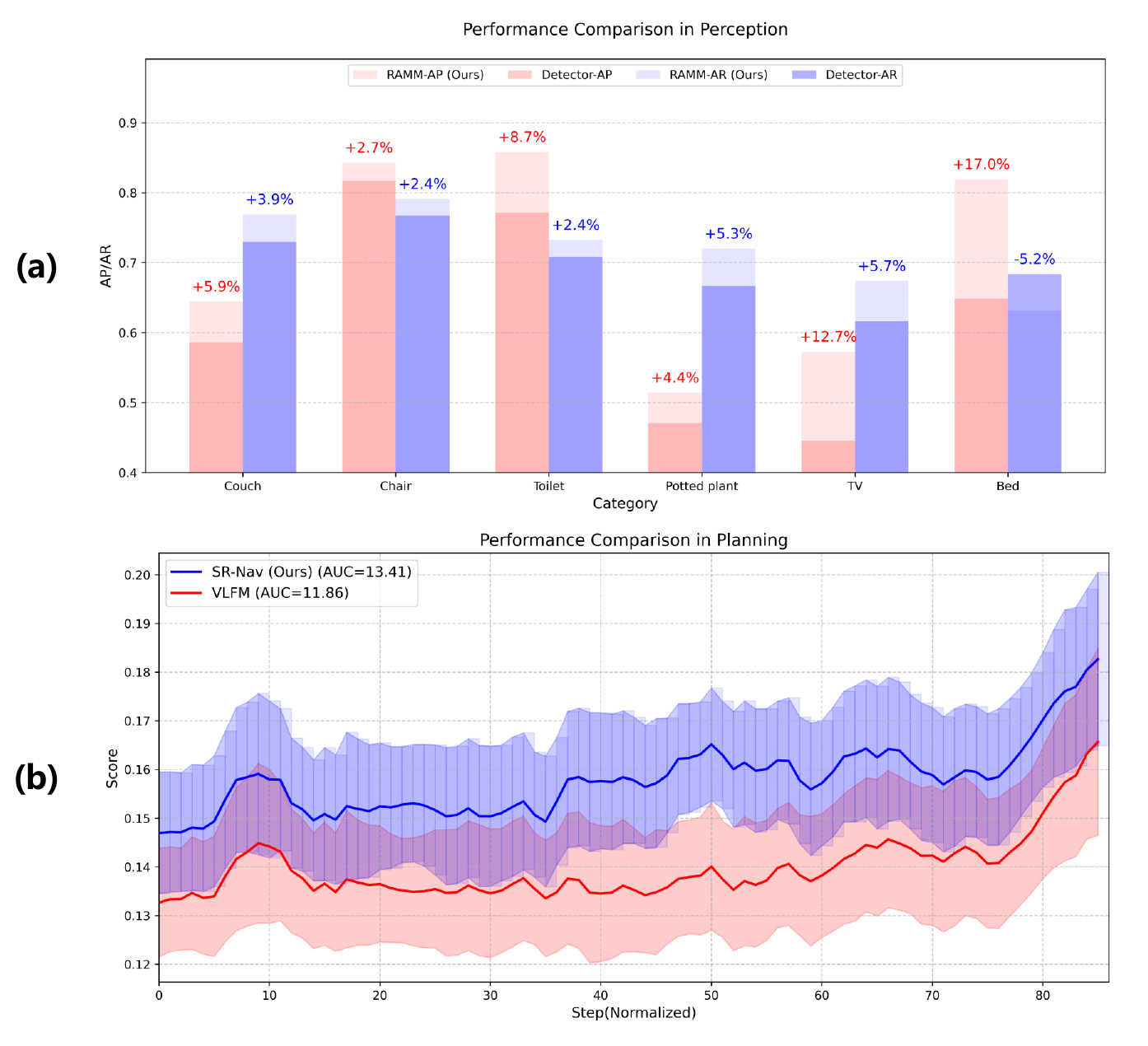} % Reduce the figure size so that it is slightly narrower than the column. Don't use precise values for figure width.This setup will avoid overfull boxes.
\label{fig:impact_visulation}
\caption{Validation of RAMM and DRPM. (a) Superior detection precision and recall across all object categories via spatial relationship matching. (b) Higher semantic similarity scores throughout planning with spatial relationship guidance.}
\end{figure}

% \begin{figure}[t]
% \centering
% \includegraphics[width=0.9\columnwidth]{images/fig8.pdf} % Reduce the figure size so that it is slightly narrower than the column. Don't use precise values for figure width.This setup will avoid overfull boxes.
% \label{fig:real_world}
% \caption{Real-world deployment. SR-Nav enables zero-shot object navigation in diverse and complex real-world environments by leveraging experience-based spatial priors to guide exploration and correct target detection results.}
% \end{figure}

\subsection{Comparison with SOTA methods}
Many recent methods construct structured scene representations to better prompt foundation models (VLMs or LLMs) for commonsense reasoning. For example, L3MVN, TopV-Nav, and Open-FM-Nav \cite{yu2023l3mvn, zhong2025_2411.16425, kuang-etal-2024-openfmnav} map observed objects onto a 2D top-down semantic map and use LLMs to reason about the best frontiers by integrating spatial and semantic information. Uni-Nav, on the other hand, builds a scene graph to encode structural relationships in the environment, which is then used to guide LLM-based frontier selection. Approaches such as GAMap, Imagine-Nav, and VLFM estimate frontier utility values to direct the agent toward the most semantically relevant frontiers for waypoint selection. 

As shown in Table {1}, we compare our method with state-of-the-art approaches \cite{yin2024sg, yokoyama2024vlfm, huang2024gamap, yin2025unigoal} on both HM3D and MP3D datasets. Empty cells indicate that the corresponding method was not evaluated on that dataset. On HM3D, our method outperforms all existing zero-shot baselines, achieving improvements of +3.8\% SR and +2.6\% SPL. On MP3D, although SR-Nav slightly underperforms Uni-Nav in SR, it surpasses it in SPL by +4.3\%, indicating superior path efficiency.

To better understand how our Spatial Relationship Guided Navigation (SR-Nav) framework facilitates navigation, we visualize representative navigation trajectories in some environments (compare with SG-Nav and VLFM). Compared to baseline methods, our approach shows a clear advantage in maintaining coherent exploration strategies, avoiding redundant loops, and focusing on high-value semantic regions, thereby achieving more interpretable and goal-directed navigation behaviors.

% \begin{figure}[t]
% \centering
% \includegraphics[width=0.9\columnwidth]{images/fig8.pdf} % Reduce the figure size so that it is slightly narrower than the column. Don't use precise values for figure width.This setup will avoid overfull boxes.
% \label{fig:real_world}
% \caption{Real-world deployment. SR-Nav enables zero-shot object navigation in diverse and complex real-world environments by leveraging experience-based spatial priors to guide exploration and correct target detection results.}
% \end{figure}

\begin{table}[ht]
\label{tab:module}
\caption{An ablation study of different modules on HM3D v0.1. 'Object' and 'Region' indicate the incorporation of object-level and region-level spatial relations during planning, respectively. 'RAMM' denotes the Relation-Aware Matching Module, which is used to verify and correct detection results.}
\centering
\footnotesize
\begin{tabular}{c|cc|c||c|c}
% \multicolumn{6}{c}\\
\toprule
\multirow{2}{*}{\textbf{Base}} & \multicolumn{2}{c|}{\textbf{DRPM}} & \multirow{2}{*}{\textbf{RAMM}} & \textbf{SR} & \textbf{SPL} \\
 & \textbf{Object} & \textbf{Region} & & (\%) & (\%) \\
\midrule
\checkmark & \xmark & \xmark & \xmark & 52.5 & 30.3 \\
\checkmark & \checkmark & \xmark & \xmark & 53.1 & 29.2 \\
\checkmark & \checkmark & \checkmark & \xmark & 54.5 & 32.3 \\
\checkmark & \checkmark & \checkmark & \checkmark & \textbf{58.3} & \textbf{33.0} \\
\bottomrule
\end{tabular}
\end{table}

\begin{table}[ht]
\caption{Performance and computational efficiency comparison.}
\label{tab:vlm_efficiency}
\centering
\setlength{\tabcolsep}{4pt} % 缩小列间距，避免表格超出边界
\begin{tabular}{ccccc}
\toprule
\textbf{Method} & \textbf{Model} & \textbf{SR(\%)} & \textbf{SPL(\%)} & \textbf{Times $\downarrow$ (min/episode)} \\
\midrule
SG-Nav & LLaMA-7B & 53.9 & 24.8 & 13 \\
\midrule
\multirow{2}{*}{WMNav} & Qwen-2.5-VL-7B & 46.1 & 20.7 & 2.0 \\
& Gemini 1.5 Pro & 58.1 & 31.2 & -- \\
\midrule
\multirow{2}{*}{ours} & InternVL-3-8B & 55.1 & 32.6 & \multirow{2}{*}{\textbf{1.5}} \\
& Qwen-2.5-VL-7B & \textbf{58.3} & \textbf{33.0} &  \\
\bottomrule
\end{tabular}
\end{table}

\begin{table*}[!t]
\caption{Prompt Sensitivity and Relationship Ablation Study.}
\centering
\begin{tabular}{c|c|c||c}
\toprule
Disturbance Type &\textbf{Prompt Content} & \textbf{SR(\%)}$\uparrow$ & \textbf{SPL(\%)}$\uparrow$ \\
\midrule
& Seems like there is a \textless\textit{target}\textgreater  ahead & 55.4 & 32.1\\
& A \textless\textit{target}\textgreater  can be in the vicinity. & 56.5 & 32.3\\
Template & Seems like a \textless\textit{target}\textgreater  ahead & 55.7 & 32.4\\
& You may find \textless\textit{target}\textgreater  nearby & 55.3 & 30.9\\
& \textless\textit{target}\textgreater & 51.3 & 28.4\\
\midrule
&W/o $R_{dis}$ & 56.0 & 31.5 \\
Relationships &w/o $R_{dir}$    & 57.3 & 32.6 \\
dropout &W/o $R_{topo}$   & 52.7 & 30.5 \\
\midrule
&\textbf{Ours}   & \textbf{58.3} & \textbf{33.0} \\
\bottomrule
\end{tabular}
\label{tab:Ablation_prompt}
\end{table*}

\begin{table*}[t]
\setlength{\tabcolsep}{4mm}
\centering
\caption{Performance comparison on object-goal navigation across six target categories and average. Best results are in \textbf{bold}. We manually verified and annotated the corresponding results. Entries based on manual annotation are marked with an asterisk~($^\ast$).}
\label{tab:detector}
\begin{tabular}{lcccccccc}
\toprule
\textbf{Method} & \textbf{bed} & \textbf{chair} & \textbf{plant} & \textbf{sofa} & \textbf{toilet} & \textbf{tv\_monitor} & \textbf{Average} \\
\midrule
L3MVN      & 52.9 & 51.6 & \textbf{46.4} & 50.1 & 41.5 & \textbf{54.2} & 49.5 \\
VLFM    & 50.1 & 66.6 & 44.0 & 48.1 & 58.5 & 34.9 & 52.5 \\
\textbf{SR-Nav (ours)}  & \textbf{55.4} & \textbf{68.9} & 42.8 & \textbf{54.1} & \textbf{65.3} & 44.8 & \textbf{58.3} \\
\textbf{SR-Nav (ours)$^{*}$}  & - & - & - & - & - & - & \textbf{61.3}$^{*}$ \\
\bottomrule
\end{tabular}
\end{table*}

\subsection{Ablation Study}
\textbf{Effect of Different Modules}. To evaluate the contribution of each module, we compare three ablated variants of our model on the HM3D v0.1 dataset. Removing DRPM-R means that during exploration, the agent is guided only by spatially associated objects, without considering the region-level associations. As shown in row 3 of Table \ref{tab:module}, the addition of the RAMM alters the conventional paradigm of relying solely on object detectors by introducing VLM-based relational matching to re-detect false positives. Table \ref{tab:module} presents a step-wise ablation analysis, where each added module consistently improves performance. Object-level and region-level relations both contribute to better scene understanding, while the RAMM module significantly boosts SR by refining target matching through relational cues.

 As shown in Fig. \ref{fig:impact_visulation}(a), that our approach boosts both average recall (AR) and average precision (AP). Although the AR for the category of ``Bed" dips slightly, the overall AR still shows a net gain. When false detections occur, our VLM-based re-detection mechanism successfully re-evaluates the scene using contextual and relational prompts, correcting both false positives (e.g., \textit{misidentified washing machine as toilet}, Fig. \ref{fig:VIS} VLFM) and false negatives (e.g., \textit{missing an occluded Toilet}, Fig. \ref{fig:VIS} SG-Nav). Additionally, our method enables the agent to efficiently explore by leveraging the structured spatial relations encoded in the DSRG. As illustrated in Fig. \ref{fig:impact_visulation}(b), we sampled 500 trajectories and computed the average semantic similarity and the 95\% confidence interval at each timestep after alignment via interpolation. As shown in Fig. \ref{fig:motivation}(c), guided by prior experience, our method acquires information that is more conducive to exploring unknown areas during the initial exploration phase. Consequently, compared to observation-only methods, it requires significantly fewer steps to achieve success in the same environments. Our DRPM module consistently achieves higher semantic relevance scores than VLFM throughout navigation, demonstrating stronger guidance cues for focused exploration in informative regions.

\begin{figure}[t]
\centering
\includegraphics[width=0.9\columnwidth]{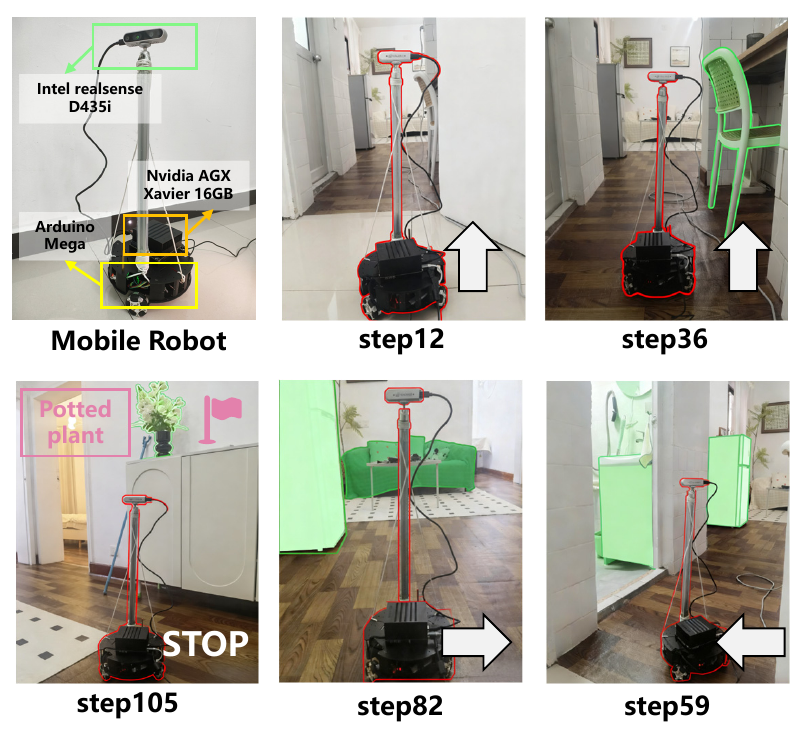} % Reduce the figure size so that it is slightly narrower than the column. Don't use precise values for figure width.This setup will avoid overfull boxes.
\label{fig:real_world}
\caption{Real-world deployment. SR-Nav enables zero-shot object navigation in diverse and complex real-world environments by leveraging experience-based spatial priors to guide exploration and correct target detection results.}
\end{figure}

\textbf{Effect of Different VLMs}. As shown in Table. \ref{tab:vlm_efficiency}, we further evaluate the impact of different VLMs on navigation performance. When no VLM is used (i.e., relying only on the static relation graph, replacing relational matching with Grounding DINO-based re-detection, and keeping all guidance cues fixed throughout navigation), our method remains competitive on the HM3D dataset. As VLMs evolve, each module in our framework benefits from enhanced reasoning capabilities, and the system as a whole has the potential to deliver significantly better performance. As shown in Table 4, our method reduces task completion time compared to other large-model-based approaches, achieving 8.7× and 1.3× improvements over SG-Nav and WmNav, respectively. Finally, we deployed our method in real-world scenarios As shown in Fig. \ref{fig:real_world}, successfully and efficiently locating the target (potted plant) in previously unseen environments.

\textbf{Effect of Different Relationships}. Table. \ref{tab:Ablation_prompt} jointly indicate that our gains are structural rather than prompt-engineering artifacts, and that explicit spatial relationships are a key contributor. Prompt sensitivity (Table. \ref{tab:Ablation_prompt}, Template) shows that replacing the natural-language template while keeping the same cues yields only small performance fluctuations, suggesting the framework is not brittle to superficial phrasing changes. In contrast, relationship ablation (Table. \ref{tab:Ablation_prompt}, Relationships dropout) causes a clear degradation: removing any single relation type consistently reduces both SR and SPL, confirming that direction, distance, and topology provide complementary supervision for both verification and planning; notably, dropping topology leads to the largest drop, highlighting its importance for long-horizon, cross-region decision making. This trend is consistent with the module ablation (Table. \ref{tab:module}): incorporating either region cues or object cues alone yields limited improvements, while combining both is better, and further adding the RAMM produces the largest gain, validating that relational verification/correction is essential when detections are noisy and the target is not directly observable. Overall, these results support our claim that performance improvements stem from relationship-grounded reasoning over structured cues, with strong robustness to reasonable prompt rewording.

\subsection{Real-World Deployment}
For real-world deployment, the proposed system was implemented on a mobile robot platform equipped with an NVIDIA Jetson AGX Xavier 16GB for onboard execution of the navigation pipeline and an Arduino Mega for low-level motion control. Scene observations were acquired using an Intel RealSense D435i RGB-D camera, which provided both visual and depth information for embodied perception. To support zero-shot target search, the onboard robot system interacted via SSH with a vision-language model (VLM) deployed on a remote server, enabling semantic scene understanding and target-oriented reasoning from the captured observations. With the integration of onboard computation, low-level control, real-time RGB-D sensing, and server-side VLM assistance, the robot was capable of performing object-goal navigation in a coherent and effective manner. In previously unseen real-world environments, the proposed method successfully and efficiently located the target object, i.e., a potted plant, demonstrating its robustness, generalizability, and practical applicability beyond simulation.

% As shown in Table \ref{tab:Ablation}, removing any relational type leads to a clear performance drop in both Success Rate (SR) and SPL. Eliminating topological relations causes the largest degradation (SR drops from 58.3\% to 52.7\%), highlighting the importance of region-level connectivity cues for navigating large and structurally constrained indoor spaces. Removing distance cues also significantly reduces SR and SPL, indicating that proximity priors are essential for efficient target search. Directional cues exhibit a moderate impact, but their removal still harms navigation consistency and leads to suboptimal path efficiency. These results collectively validate that all three relational channels are complementary and jointly necessary for robust VLM-guided object navigation.

\section{Conclusion}
\label{sec:Conclusion}

In this work, we propose SR-Nav, a novel framework that leverages structured spatial relationships to enhance both perception and planning for zero-shot object navigation. By constructing a Dynamic Spatial Relation Graph (DSRG) and integrating vision–language models, SR-Nav achieves two key capabilities: relation-aware matching, which corrects unreliable or missing detections through spatial priors, and relation-guided planning, which enables more efficient and targeted exploration in previously unseen environments. Extensive experiments on HM3D and MP3D demonstrate that SR-Nav consistently outperforms existing state-of-the-art methods in both Success Rate (SR) and Success weighted by Path Length (SPL). 

In future work, we will explore extending this approach to multiple agent platforms, such as quadruped robots or mobile robots, to validate its generality. We will also investigate how to leverage sparser and longer-range spatial relationships in outdoor environments to support navigation.

{
    \bibliographystyle{IEEEtran}
    \bibliography{ref}
}

\newpage

 

% \end{IEEEbiography}



\vfill

\end{document}